\documentclass[conference,compsoc]{IEEEtran}
\ifCLASSOPTIONcompsoc
  \usepackage[nocompress]{cite}
\else
  \usepackage{cite}
\fi
\raggedright
\usepackage[justification=justified, singlelinecheck=false]{caption}  
\usepackage{float}
\usepackage{lineno}
\usepackage[utf8]{inputenc}

\usepackage{array}
\usepackage{algorithm}
\usepackage{algpseudocode}

\usepackage{graphicx}
\usepackage{multirow}
\usepackage{amsmath}
\usepackage{ragged2e} 
\usepackage{array} 
\usepackage[table]{xcolor} 
\usepackage{lipsum} 

\usepackage{tabularx} 
\usepackage{booktabs} 
\usepackage{makecell} 
\usepackage{csquotes}
\usepackage{graphicx}
\usepackage{multirow}
\usepackage{amsmath}
\usepackage{subfigure}
\usepackage{multimedia}

\ifCLASSINFOpdf
\else
\fi
\begin{document}
%
\title{Inherent Diverse Redundant Safety Mechanisms for AI-based Software Elements in Automotive Applications}

\author{\IEEEauthorblockN{Mandar PITALE }
\IEEEauthorblockA{NVIDIA\\
Santa Clara, CA\\
Email: 	mpitale@nvidia.com}
\and
\IEEEauthorblockN{ Alireza ABBASPOUR}
\IEEEauthorblockA{Qualcomm\\
San Diego, CA\\
Email: aabbaspo@qti.qualcomm.com}
\and
\IEEEauthorblockN{Devesh UPADHYAY}
\IEEEauthorblockA{Saab\\
Canton, MI\\
Email: deveshu@gmail.com}
\\
}


%

\pagestyle{plain} 


\maketitle
\justifying 
\begin{abstract}  
  This paper explores the role and challenges of Artificial Intelligence (AI) algorithms, specifically AI-based software elements, in autonomous driving systems. These AI systems are fundamental in executing real-time critical functions in complex and high-dimensional environments. They handle vital tasks like multi-modal perception, cognition, and decision-making tasks such as motion planning, lane keeping, and emergency braking. A primary concern relates to the ability (and necessity) of AI models to generalize beyond their initial training data. This generalization issue becomes evident in real-time scenarios, where models frequently encounter inputs not represented in their training or validation data. In such cases, AI systems must still function effectively despite facing distributional or domain shifts. This paper investigates the risk associated with overconfident AI models in safety-critical applications like autonomous driving. To mitigate these risks, methods for training AI models that help maintain performance without overconfidence are proposed. This involves implementing certainty reporting architectures and ensuring diverse training data. While various distribution-based methods exist to provide safety mechanisms for AI models, there is a noted lack of systematic assessment of these methods, especially in the context of safety-critical automotive applications. Many methods in the literature do not adapt well to the quick response times required in safety-critical edge applications. This paper reviews these methods, discusses their suitability for safety-critical applications, and highlights their strengths and limitations. The paper also proposes potential improvements to enhance the safety and reliability of AI algorithms in autonomous vehicles in the context of rapid and accurate decision-making processes.    
\end{abstract}


%
\IEEEpeerreviewmaketitle

\section{ Introduction}
In the rapidly evolving field of artificial intelligence, the issue of model overconfidence has emerged as a pressing concern and has attracted the attention of researchers, developers, and the wider public. This paper delves into the problem of AI models displaying overconfidence, a phenomenon with significant implications for safety and reliability. Overly confident AI models often make predictions with unjustified certainty, frequently failing to acknowledge the inherent uncertainty in complex real-world scenarios \cite{rabanser2019failing}. For example, overconfident perception models in autonomous driving applications may inaccurately classify objects on the road, erroneously projecting unwarranted certainty in object detection and the ensuing action \cite{melotti2022reducing}. Likewise, overconfident path prediction models can make incorrect assumptions about the actions of other vehicles, potentially resulting in or creating hazardous interactions \cite{yoon2020road}. Furthermore, overconfident path planning algorithms might underestimate the risks associated with specific maneuvers, endangering not only the ego vehicle and its occupants but also the surrounding traffic environment \cite{kahn2017uncertainty}. These overconfident behaviors can lead to severe consequences, particularly in domains where AI systems hold responsibility for critical decision-making. 

To address these potential risks, this paper aims to investigate the root causes of overconfidence in AI models, analyze potential consequences, and, significantly, chart a course toward innovative solutions that can enhance the safety and dependability of AI systems. In doing so, we will examine existing mitigation techniques while also advocating for a proactive approach that encourages interdisciplinary collaboration between fields of AI, system safety, and reliability to address this emerging challenge in the field of artificial intelligence.
Various methods and solutions have been introduced to solve the problem of uncertainty and overconfidence in AI models. To name a few, in-distribution detectors, out-of-distribution detectors \cite{serra2019input}, certainty estimators \cite{gal2016dropout}, distribution shift detectors \cite{rabanser2019failing}, and adversarial perturbation detectors \cite{mustafa2019adversarial}. We assess these methods and propose a "diverse, redundant safety mechanism" as an alternate approach for consideration. The diverse, redundant safety mechanism, when integrated into the AI element architecture, harnesses the combined strengths of conventional methods to mitigate their respective limitations effectively.  
Introducing a majority voter or plausibility checker becomes essential to employ multiple redundant safety measures to collectively determine the state of the input. Such a voter likely combines the benefits of different error detection techniques/measures that impact the AI software development process and AI architecture to determine if the data is unseen and avoid wrong AI element output. This checker and the overarching methodology are designed to enhance critical metrics, including reducing FP, FN, etc.

The main contributions of this paper can be summarized as follows: 1) Reviewing the state of art literature related to AI confidence. 2) Summarizing the advantages and disadvantages of these methods with an automotive safety lens. 3) Proposing new solutions to enhance the safety of AI models through diverse redundant safety architectures and analyzing the broader impact of the safety measures on the overall reliability of an AI element.

The remainder of this paper is organized as follows. Section 2 discusses the problem of overconfident AI models and uncertainties in model behavior against unpredicted scenarios. Section 3 provides definitions that are critical for understanding the solutions proposed. Section 4 reviews the state-of-the-art solutions for managing overconfident AI models and discusses the concept of safety mechanisms for AI-based SW elements based on those solutions. Section 5 provides our proposed solution to enhance the safety of the AI model using inherent diverse redundant architecture. Conclusion and future work are presented in Section 6.

\section{Overconfident AI Models}

While extending the generalization capabilities of AI models remains a topic of research pursuit, one of the persistent challenges that researchers and practitioners face is dealing with overconfident AI models. Overconfident models significantly impact the reliability and trustworthiness of AI-generated outputs and of any downstream functions reliant on these outputs. \textit{ For this paper, we define an overconfident model as one that produces predictions with high certainty, even in situations where the inherent uncertainty is substantial.  In other words, an overconfident model overestimates its own accuracy and reliability, especially when generalizing}.  
One can easily derive a representation of an overconfident model as follows. 
Consider a population of size \(P + N\) with \(P = \) total number of positive examples and \(N = \) total number of negative examples.  
Then, for an overconfident model, we can say:
\begin{equation}
\frac{TP+FP}{P+N} > \frac{P}{P+N}
\end{equation}
where  \(TP, FP\) are the models True and False Positive predictions, respectively.  Specifically, an overconfident model satisfies these two conditions:  
(\(TP+FP) > P\) and \(TP < P\).  
Another simplistic naive model for an overconfident predictor may be written as: 
\begin{equation}
\mbox{ Model Confidence }  = 1 + \frac{(Model_{acc} - True_{acc})}{(Model_{acc} + True_{acc})}
\end{equation}

where \(acc\) = accuracy. This simple definition shows that for \(Model_{acc} > True_{acc}\) we get a confidence score \(> 1\) .

Overconfident models lead to potentially erroneous decisions and, hence, detrimental consequences in critical applications. This phenomenon has raised serious concerns within the AI community, prompting researchers to delve into the intricacies of this problem.

The impact of overconfident AI models can be far-reaching, affecting decision-making processes across various domains, particularly autonomous systems. In autonomous driving, for instance, overconfident AI diagnostic models might exhibit unwarranted certainty in their path predictions, potentially leading to incorrect path planning and leading the vehicle to another vehicle path and causing an accident. 
The fatal incidents involving autonomous vehicles in Arizona \cite{ntsb-uber-2018} and Florida \cite{ntsb-tesla-2016} highlight a critical issue in AI-operated systems: overconfidence in decision-making models. This overconfidence is evident when these systems, trained and validated on limited scenarios, encounter real-world complexities that deviate from their training data, leading to miscalculations or complete oversights.

In the Arizona incident involving an Uber autonomous vehicle, the system's predictive model exhibited overconfidence in its ability to discern and react to complex scenarios \cite{ntsb-uber-2018}. The perception model failed to identify the pedestrian accurately and did not engage in timely corrective action, suggesting a disconnection between the AI's predictive certainty and the objective ambiguity of real-world data. This overconfidence, rooted in the incongruence between the training data and the nuanced variables of live environments, highlights the susceptibility of AI systems to overlook or misinterpret unforeseen inputs, thereby leading to catastrophic outcomes.
Similarly, the Tesla accident in Florida manifests another dimension of overconfidence in AI models. The vehicle's autonomous system was not equipped to recognize the white semi-trailer against a bright sky, a phenomenon indicative of perceptual aliasing. Despite this, the system did not significantly lower its confidence or engage precautionary measures, continuing instead under the presumption of precise navigation \cite{ntsb-tesla-2016}. This incident accentuates that overconfident AI may not only misinterpret data but might also entirely miss critical information, all while maintaining a high confidence level in its flawed perception and the ensuing decision-making processes.

The causal cascade of prediction/sensing error on downstream actions (decisions) is well studied in Control theory. In non-AI systems, errors are typically propagated as sensor errors. In AI systems, the sensing equivalent is the model prediction error.  Thus, given a decision or action \(A\)
\begin{equation}
A = F(X_{i})
\end{equation}
for an \(i\) dimensional state \(X\) and some state to action mapping \(F\), it is easy to represent the impact of state prediction error \(\delta X\) on \(A\) as:
\begin{equation}
\delta A = \sum_{i} (\frac{\delta F}{\delta X_{i}}\Delta X_{i}) 
\end{equation}
where the term 
\(\frac{\delta F}{\delta X_{i}}\) is the sensitivity of the action function \(F\) to the state \(X_{i}\) and \(\delta X_{i}\) is the error in prediction of the state \(X_{i}\).  This simple affine error cascade model shows the impact of error aggregation, especially for high dimensional systems, on the AI-driven action \(A\). Critically, argue that the overconfident model will "fail silently" and will not provide adequate warning to the decision system to counter the accruing error (debt).
This false sense of security from overconfident AI models can lead to hazardous driving decisions in complex real-world scenarios. For example, overconfident AI models can fail to adequately assess risky situations, leading to accidents or misinterpretations of critical road features \cite{bojarski2016end}. Moreover, in dynamic environments where unexpected events frequently occur, such as sudden obstacles or adverse weather conditions, overconfident AI models alone may not exhibit the necessary adaptability, thus jeopardizing the safety of passengers and other road users. Hence it is critical to not only understand the root causes of these model behaviors but also put in place measures that can provide safety guarantees for mission critical systems.

The root causes of the problem of overconfident AI models can be attributed to various factors, including but not limited to biased training data, sub-optimal model architectures, and inadequate uncertainty quantification techniques. Biased training data, stemming from historical data that may not accurately represent the diverse range of real-world scenarios, can inadvertently influence AI models to exhibit unwarranted confidence in their predictions \cite{raji2019actionable}. Furthermore, sub-optimal model architectures, lacking representation capacity and, therefore,  lacking in their ability to capture complex patterns and nuances within the data, can contribute to the propagation of overconfident behaviors in AI systems \cite{melis2017state}. Additionally, the lack of robust uncertainty quantification techniques, which are essential for estimating and communicating the uncertainties associated with AI predictions, further exacerbates the issue of overconfidence \cite{kendall2018multi}. Addressing these root causes requires a comprehensive approach that involves meticulous data curation, sophisticated model design, and the integration of effective uncertainty estimation methodologies within AI frameworks. This is especially crucial when considering the safety of AI-enabled systems.    

\section{Definitions}
In this section, we provide definitions that are important for providing a sound basis for understanding the proposed solution. The first two definitions are from state-of-the-art functional safety standards for automotive, i.e. ISO 26262:2018. The later definitions are our own contributions.

\textit{a. Safety mechanism: technical solution implemented by E/E functions or elements, or by other technologies, to detect and mitigate or tolerate faults or control or avoid failures in order to maintain intended functionality or achieve or maintain a safe state.
}

\textit{b. Safety measure: activity or technical solution to avoid or control systematic failures and to detect or control random hardware failures or mitigate their harmful effects.
}

As such, a safety mechanism is a subset of safety measures. Therefore,  safety mechanisms are typically run-time measures, and safety measures can be run-time or design-time measures \cite{guerin2023out}.

\textit{c. Run-time measure: An entity or artifact that is evidently either present or its effects are perceivable during deployment of the system.
}

\textit{d. Design-time measure: A process/task step such as verification.
}

\textit{e. Inherent safety mechanism:  An entity that is embedded or an inseparable part of the AI-based SW element.}

The safety mechanisms for AI-based SW elements can be embedded or inseparable parts of those SW elements or outside the SW elements but part of the AI system that contains those AI elements \cite{ferreira2021benchmarking}. In this paper, the former are referred to as Inherent safety mechanisms; they are the primary focus of the paper as they can be essentially used to perform a voting function when more than one such inherent safety mechanism is available. However, while doing the research and investigation for each of the error detection methods, if it is identified that some measures are design time or if part of the AI system or if part of non AI-based software elements, then those will be identified that they can't be used for such a voting function.

\section{ Related works}

The full scope of an AI-based software element includes the model architecture, parameters, and the training process. Given the probabilistic nature of AI elements, their behavior with respect to the generated output also exhibits structured probabilistic behavior that can sometimes be the root cause of model prediction errors. The root causes of these errors depend on the inherent nature and properties of the AI-based software elements. As such, these errors are categorized based on the different error types as identified by Mohseni et al. \cite{mohseni2019practical} and discussed in detail in a subsequent sub-section.  It is likely that these categorizations have some overlap and synergies, and the detection of these error types might also have certain commonalities. OOD Detection, In-Distribution Detection, and Distribution shift monitoring are discussed due to the open-world nature of the context/domain where AI elements are required to be used. This simply means OOD or In-Distribution etc. are error types where as OOD detection and In-Distribution detection are methods to detect the respective errors.
Uncertainty Estimation is discussed based on the fundamental way that the AI elements are built based on probabilistic and/or stochastic methodologies. Adversarial perturbation is chosen as it is common for AI-based elements to exploit its vulnerability and limitation, leading to errors in their outputs.

\subsection{Inherent safety measures/mechanisms for AI-based SW elements and their pros-cons}
In the subsections of this section, first, we provide the context of each method as stated in the context/introduction section, and then we list the several measures that help to fulfill that method. The measures are listed/chosen based on the assumption/hypothesis that it is likely a run-time measure and can be used as a part of a voter. We also list a generic formula or expression based on our research for each of the measures to support if the diversity claims can be made. The idea here is that if 2 or more detection measures for the same or different error detection method that provides comparable output, then these measures can possibly foster diversity as they are based on different mathematical expressions.

\subsubsection{Out-of-distribution detection}
AI-based software elements are trained based on a large set of data. The out-of-distribution (OOD) data is the data that is outside the normally considered distribution. It can also be outside the decision boundary of the considered distribution but not too far from it, or it is remarkably different than the considered distribution. If OOD data is detected incorrectly to be part of the distribution and/or lately, it can possibly lead to wrong decisions, leading to hazards.

Here, a generic pseudo algorithm for OOD is presented at Algorithm 1. This algorithm is designed to categorize samples in a given data-set as either In-Distribution (ID) or Out-of-Distribution (OOD). Given a batch of mixed training labeled samples and unlabeled samples \cite{mohseni2020self}, along with a set of predefined OOD classes \( k \) and a learning coefficient \( \lambda \), the algorithm aims to distinguish between ID and OOD samples effectively. 
It initializes a model with training data and then iteratively processes each sample in the batch. For each sample, it computes the probability of its being in distribution. If this probability is low, the sample is considered potentially OOD, and an OOD score is computed using \( \lambda \). The algorithm then evaluates this sample against each OOD class, assigning it to the class with the highest calculated probability. Samples with a high in-distribution probability are categorized as ID. This method allows for a systematic identification of OOD samples and their classification into relevant categories.


\begin{algorithm}
\caption{Out-of-Distribution (OOD) Detection}
\begin{algorithmic}[1]

\Require Batch of mixed samples $S$, OOD classes $k$, learning coefficient $\lambda$
\Ensure ID/OOD categorization for each sample, OOD class if applicable
\State Initialize model $M$ with training data
\For{each sample $s \in S$}
    \State Compute $p_{\text{ID}}(s)$ using $M$
    \If {$p_{\text{ID}}(s)$ is low}
        \State $s$ is potentially OOD
        \State Compute OOD score using $\lambda$
        \For{each class $c \in k$}
            \State Compute $p_c(s)$
        \EndFor
        \State Assign OOD class to $s$ with highest $p_c(s)$
    \Else
        \State $s$ is ID
    \EndIf
\EndFor

\end{algorithmic}
\end{algorithm}

\textbf{Softmax function}: The softmax function is primarily used towards the output layers of the AI-based Software elements (e.g. DNN) that generate a probability distribution over multiple classes of input data. The outcome of the softmax may be likely incorrect or misleading, although it indicates higher confidence for some classes  \cite{cheng2023runtime}. This could be because OOD data is not seen during the training or due to overconfidence in prediction. The overconfidence in prediction can be due to the lack of confidence in calibration techniques during training and/or due to the absence of specific resiliency measures. 

\begin{equation}
  \sigma(\vec{x})_i :\stackrel{\text{def}}{=} \frac{e^{\vec{x_i}}}{\sum_{i=1}^{d_L}{\vec{e^{x_i}}}}
\end{equation}

\textit{As softmax is the fundamental building block in AI-based SW classifier elements and typically without specific additional measures (such as reject class(d)), it can't be used as one of the redundant safety mechanisms by itself. }

\textbf{Temperature scaling}: Temperature scaling is used to improve prediction confidence such that AI elements can make better classifications by adding the scaling factor. It often works in conjunction with the softmax function. The optimum value of the scaling factor needs to be chosen such that there is clear segregation between the in-distribution and OOD input data. In addition, the small perturbation is also added at times to increase the softmax score for the in-distribution data \cite{liang2017enhancing}. 

\begin{equation}
  \sigma(\vec{x}, T)_i :\stackrel{\text{def}}{=} \frac{e^{{\vec{x_i}}/T}}{\sum_{i=1}^{d_L}{e^{{\vec{x_i}}/T}}}
\end{equation}

\textit{Typically for temperature scaling, there is no necessity to change the training process or the AI element network architecture. As temperature scaling is used in the context of softmax, it can't be used as one of the redundant safety mechanisms by itself.  }

\textbf{Mahalanobis distance}: Mahalanobis distance \cite{NEURIPS2018_abdeb6f5} is a confidence score based on an induced generative classifier under Gaussian discriminant analysis (GDA). The confidence score is reflected in the softmax function towards the output layers of AI elements-based classifiers. As such the Mahalanobis distance measure can be viewed as similar to temperature scaling from the AI model architecture perspective. 

\begin{equation}
  z :\stackrel{\text{def}}{=} \frac{f_{i}^{1\rightarrow l} (\vec{in})}{\sigma}
\end{equation}

For Mahalanobis distance, no changes in the network architecture are necessary. This method can detect OOD samples, adversarial attacks, and class incremental learning. This method was also found more robust in scenarios where the training dataset has some noisy, random labels or a small number of data samples. 

\textit{As Mahalanobis distance is used in the context of softmax, it can't be used as one of the redundant safety mechanisms by itself. }

\textbf{Isolation forest}: Isolation forest is an outlier detection technique where multiple decision trees are used to isolate the anomalies from in-distribution data; as such, it is an ensemble method \cite{liu2008isolation}.  The isolation Forest algorithm randomly selects a feature and then randomly selects a split value between the maximum and minimum values of the selected feature. The nature of this technique is recursive, such that the maximum height of a single tree is reached. The OOD data leads to having short average path lengths in isolation forests; as such, they can be identified more quickly than in distribution data. The isolation forest can be applied to input data directly or to the intermediate layers in case of a DNN {\cite{luan2021out}}.

\begin{equation}
  s(x_p) =2^{\frac{E\big(h(x_p)\big)}{c(n)}} ,~~ c(n)=2H(n-1)-(2(n-1)/n)
\end{equation}

Given that  $h_{x_p}$ denotes the path length for an individual tree corresponding to the input $x_p$, and $E\big(h(x_p)\big)$ represents the average of path length across all trees; the harmonic number $H(i)$ can be approximated using $H(i) = \ln(i) + \gamma$, where $\gamma$ is Euler's constant, approximately 0.5772. A sample is more likely to be an outlier if it has a greater anomaly score.

\textbf{Local outlier factor}: LOF is a detection technique to identify the outliers/anomalies. It is an unsupervised method that computes the local density of a given input and compares it with the local density of its k nearest neighbors. This method is different than the global outlier detector method in the fact that LOF primarily works on the local density as the global outlier detector works on the global characteristics of the entire considered dataset. Local density is calculated based on the inverse of average reachability distance with reference to the local neighbors. The samples with much lower density than their neighbors are considered outliers \cite{luan2021out}.

\begin{equation}
  s(x_p) = LOF_k(x_p) = \frac{1}{|N_k(x_p)|} \sum_{q \in N_k(x_p)} \frac{lrd_k(x_q)}{lrd_k(x_p)}  
\end{equation}

where $LOF_k(x_p)< 1$ means that  $x_p$ has a higher density than its neighbors (likely to be an inlier); $LOF_k(x_p)> 1$ means $x_p$ has a lower density than its neighbors (likely to be an outlier).

Luan et al. proposed the run-time monitor based on Isolation forest and Local Outlier Factor \cite{luan2021out}. This run-time monitor for OOD detection (also referred to as OOD detector) works by tapping one or more hidden layers of the DNN and by applying the isolation forest to the neuron activations of one or more hidden layers. The number of OOD detectors is dependent on the number of hidden layers to be monitored. If at least one of the OOD detects that the input data is OOD, then the input data is determined to be OOD. For optimization of compute time, typically, the OOD detectors are traversed from right to left as there are fewer neurons towards the right side of the DNN.

\textit{The run-time monitor based on the Isolation forest and Local Outlier Factor provides an output to indicate if the data belongs to in-distribution or OOD; as such, this output can be used as one of the inputs of the voter.}

\textbf{Reject Classes}: Reject class or classes are typically added at the output layers of the AI element to indicate that the specific input was not likely seen and its features are not sufficiently detected during the training process \cite{NIPS2017_4a8423d5}. These inputs cannot be too far from the decision boundary or likely be too far based on the type of task or the operation design domain. When the input with the lower confidence score is encountered, then it is identified as belonging to the "Reject" class. Mohseni et al. \cite{mohseni2020self} have proposed multiple reject classes with different OOD detection scores, such as weighting softmax, the entropy of softmax vector, and a combination of weighted softmax, etc. Here, the self-supervised learning approach is used in 2 steps: in the first step, the labeled data is used, and in the second step, a combination of labeled and unlabeled data is used. In addition to rejecting the OOD inputs, the method also performs quite well toward the inputs that are near the boundaries of training distribution.

Introducing a reject class in a classifier implies adding a third probability P(reject); thus, in this setting, the revised probabilities are $ P(positive) + P(negative) + P(reject) = 1$. Now, with $3$ classes, we can write a compact probability of classification as: 
\begin{equation}
  P(\text{class } i | \mathbf{x}) = \frac{e^{z_i}}{\sum_{j=1}^{3} e^{z_j}}
\end{equation}
where ${z_j}$ is the logit corresponding to class $i$\\.

\textit{As the reject class is an additional output, it can definitely be used as an input for voters.
}

\subsection{Uncertainty Estimation}
Uncertainty in model prediction can be due to epistemic uncertainty and aleatoric uncertainty. Epistemic uncertainty also known as model/knowledge uncertainty \cite{abdar2021review} is due to insufficient knowledge/data or insufficiency of required feature representation. It can be addressed by bringing more relevant data, improving the model, etc. Aleatoric uncertainty can be reduced due to variability in data due to noise, measurement errors, etc. 

\textbf{Monte Carlo Dropout}: Certain layers of the AI element dropped during each forward path such that there are multiple passes, and the uncertainty is reflected in the softmax output. To obtain a dependable uncertainty estimate, tens or hundreds of forward passes are necessary \cite{luan2021out}.

Expressed mathematically, we can represent Monte Carlo Dropout as follows: for a given test input $x_{test}$ a NN model with parameters \(\theta\) will produce an output $y_{test}$ with a distribution over $N_{s}$ samples as:

\begin{equation}
P(y_{\text{test}} | \mathbf{x}_{\text{test}}, \theta, \mathcal{D}) \approx \frac{1}{N_{s}} \sum_{t=1}^{N_{s}} P(y_{\text{test}} | \mathbf{x}_{\text{test}}, \theta_t)
\end{equation}

\textit{There is no additional output is produced in the process normally, as such this measure cannot be used as one of the inputs of the voter and additional post-processing may be necessary.}

\textbf{Deep Ensemble}: It involves multiple models with the same/different architecture trained with same/different initialization parameters using same/different data such as slightly perturbed data or augmented data or a dataset split into multiple sub-sets \cite{mohammed2023comprehensive}. The ensembles can be sequential or parallel and homogeneous/heterogeneous based on design. Typically, towards the end of the deep ensemble, multiple prediction outputs are combined into one output, likely based on different methods such as bagging, boosting, stacking, averaging, etc.  Each of the methods impacts the performance indicators and computing power. The formula for weighted average voting is given by:

\begin{equation}
y* =  \frac{\sum_{j=1}^{m} {W_j}{X_i}}{\sum_{j=1}^{m} {W_j}}
\end{equation}

where $W_j$ is the weighted average for each class,  $X_i$ is the confidence prediction of each classifier, and $m$ is the total number of classes.

\textit{As deep ensemble aggregates output of more than one AI model, it doesn't generate additional diagnostics output (such as reject class or ID/OOD) as such it can't be used as input of a voter for and additional post-processing may be necessary.}

\textbf{Baysian NN}: Bayesian neural networks are stochastic neural networks trained using the Bayesian and reflect uncertainty in their predictions. Instead of representing the weights, activations and outputs using single point estimates they represent them with the probability distribution \cite{9756596}.

To make predictions for a new input  $x_{new}$ one needs to compute the predictive distribution for the target variable, $y_{new}$ by marginalizing over the parameters:
\begin{equation}
    P(y_{\text{new}} | \mathbf{x}_{\text{new}}, \mathcal{D}) = \int P(y_{\text{new}} | \mathbf{x}_{\text{new}}, \boldsymbol{\theta}) \cdot P(\boldsymbol{\theta} | \mathcal{D}) \, d\boldsymbol{\theta}
\end{equation}\\  
where $\mathcal{D}$ represents the observed data. 

\textit{The Bayesian NN reflects the uncertainty in the model output with probabilistic distribution; additional post-processing may be necessary to be used as an input of voter.}

\subsection{In-distribution Detection}
In-distribution (ID) data is the opposite of the OOD data such that it is part of the considered distribution or similar but can still lead to errors such as misdetection, misclassification because the data may be towards the edge of the decision boundary or not sufficiently represented during training. 

\subsection{Distribution Shift Detection}
Distribution shift detectors play a crucial role in AI safety, particularly in countering the overconfidence issue often inherent in AI predictions. These mechanisms identify instances of significant deviation in a model's input from its original training distribution, a critical feature for real-time applications where data dynamics can change unpredictably. The strength of these detectors lies in their ability to enhance model robustness and maintain high-reliability levels by signaling the need for potential corrective measures like model re-adaptation or retraining. However, they are not without their shortcomings. These systems can be computationally expensive and inherently complex, especially when dealing with high-dimensional data or when a detailed understanding of the initial training distribution is lacking. Also, the effectiveness of distribution shift detectors relies heavily on the comprehensiveness of the model's training phase, which might not fully encompass the multifaceted nature of real-world data variability \cite{rabanser2019failing}.\\
Covariate shift, label shift, and concept drift are three fundamental distribution shifts encountered in AI training \cite{huyen2022designing}:  Covariate shift occurs when the distribution of input data changes between training and testing phases, denoted as \( P_{\text{train}}(X) \neq P_{\text{test}}(X) \), while the conditional distribution of output given input remains constant (\( P_{\text{train}}(Y|X) = P_{\text{test}}(Y|X) \)). Label shift happens when the distribution of output labels alters (\( P_{\text{train}}(Y) \neq P_{\text{test}}(Y) \)), but the conditional distribution of inputs given the label stays the same (\( P_{\text{train}}(X|Y) = P_{\text{test}}(X|Y) \)). Concept drift is a change in the joint distribution of inputs and outputs over time (\( P_{\text{train}}(X, Y) \neq P_{\text{test}}(X, Y) \)), implying a modification in the underlying process generating the data. These shifts are critical for understanding and designing resilient AI systems that can adapt to real-world data changes.

Distribution shift detection in AI models generally involves identifying when the data distribution at inference differs significantly from the training data distribution. A common approach to quantify this shift is to use statistical distance measures. The Kullback-Leibler (KL) divergence is often used for this purpose \cite{sugiyama2012machine}. The formula for KL divergence is given by:
\begin{equation}
D_{KL}(P || Q) = \sum_{x \in \mathcal{X}} P(x) \log\left(\frac{P(x)}{Q(x)}\right)
\end{equation}
where $P$ is the probability distribution of the training data, $Q$ is the probability distribution of the new data, and $X$ is the set of all possible events. To detect distribution shifts, one would compute the KL divergence between the training distribution and the distribution observed during operation. A significant divergence would indicate a distribution shift.\\


To address distribution shifts, Rabanser et al. \cite{rabanser2019failing} outline three primary objectives when dealing with distribution shifts: (i) early detection of such shifts, (ii) characterizing the nature of the shift, particularly in test samples, and (iii) assessing the harm posed by the shift. Their work explores this through statistical two-sample testing, comparing the source and target data distributions. However, there are challenges in extending traditional hypothesis testing to high-dimensional data, such as images, as existing kernel-based methods suffer scalability issues and reduced statistical power \cite{ramdas2015decreasing}.  Kulinski et al. \cite{kulinski2020feature} extend the distribution shift detection to localizing the cause of distribution shift to a specific set of features or a specific set of sensors.  Nonetheless, both the methods proposed by \cite{ramdas2015decreasing} and \cite{kulinski2020feature} are based on the hypothesis testing principle, which refers to the statistical analysis used during the model development and evaluation. Detection of distribution shifts does not directly alter the AI model's architecture. Instead, these detectors serve as a monitoring tool, signaling when a model may not perform optimally due to data changes. This detection can then lead to various actions that might affect the model, either through retraining, adaptation, or other modifications \cite{kulinski2020feature}.

\textit{In summary, while the above discussed distribution shift detectors play a critical role in signaling potential issues with model performance, they do not provide an additional diagnostic output, hence it cant be used as a part of a voter as is. Their primary function is to alert when a model might need retraining or other adjustments to align with the new data distribution. The actual response to a detected shift, whether it involves retraining, model adaptation, or other adjustments, depends on the specific requirements and capabilities of the system.}

\subsection{Adversarial Perturbation Detection}

Adversarial perturbation detectors serve as a vital defense mechanism in AI systems, ensuring these models maintain integrity and performance consistency in the face of adversarial attacks. These attacks, characterized by subtle input manipulations intended to induce model errors, pose a significant threat to AI's reliability, especially in security-sensitive applications. Adversarial detectors identify inputs intentionally crafted to exploit model vulnerabilities, thereby safeguarding system performance and reliability. Despite their importance, these detectors face considerable challenges, including the difficulty in discerning sophisticated adversarial inputs from legitimate ones, potentially leading to false positives. Additionally, they must contend with the adaptive nature of adversaries who are continually refining their attack methodologies. Nonetheless, implementing these detectors is crucial for instilling trust in AI systems and is an active area of research aiming to balance efficacy with efficiency \cite{metzen2017detecting}.

Klinger et al. introduced an approach to adversarial perturbation detection based on edge extraction from various outputs, such as semantic segmentation and depth estimation \cite{klingner2022detecting}. They employed edge consistency as a loss function and a metric for detecting adversarial perturbations. Their work also outlines specific contributions, including the proposal of a detection method and a consistency loss function, as well as validation against sophisticated attacks like the Orthogonal-PGD (O-PGD) attack \cite{bryniarski2021evading}.

Goel et al. presented a method called the locally-optimal generalized likelihood ratio test (LO-GLRT) designed for detecting targeted attacks on classifiers involving universal adversarial perturbations (UAPs) \cite{goel2022fast}. They offer both analytical and empirical validation of the LO-GLRT, revealing its detection probability closely matches the theoretical lower bound. The method employs a Gaussian distribution as a surrogate for the usually unknown input distribution, making it mathematically tractable and computationally efficient. Importantly, LO-GLRT is shown to outperform the existing perturbation rectifying network (PRN) detector \cite{akhtar2018defense} in both statistical and computational terms, boasting a running time of at least 100 times faster. The variational approach for LO-GLRT in \cite{goel2022fast} is presented by the following equation:

\begin{equation}
    U_n(x) = \max_{t \in T} \sum_{i=1}^{n} h_{i,t}^T \tilde{\Sigma^{-1}} (x_i - \tilde{\mu})
\end{equation}
where \(\tilde{\mu}\) and \(\tilde{\Sigma}\) represent the mean vector and covariance matrix of the common distribution across the input subvectors, respectively. To achieve a consistently high detection probability across perturbation strengths within the interval \([0, \epsilon^*]\), for a given sufficiently large \(\epsilon^*\),  \(\psi^*\) is introduced that address the subsequent minimization problem:
\begin{equation}
    \psi^*(\rho, \nu) = \underset{\psi \in \Psi}{\mathrm{argmax}} \int_{0}^{\epsilon^*} \hat{P}_{D,l}(\epsilon, \rho; \nu, \delta_\psi) d\epsilon
\end{equation}
\textit{As discussed above, adversarial perturbation detectors work by monitoring the inputs and outputs of a model in run-time to identify anomalies or patterns that suggest the presence of adversarial attacks. However as it is closed box approach that means it likely alters the AI system architecture, it doesn't not provide an additional diagnostic output which can be used a part of voter.}

Table.\ref{Tab:OOD} further summarizes the methods and measures that are explained in this section before their pros and cons, whether the specific measure is a run-time measure if it impacts the AI element architectural modification such that it can be used as an input to a voter. 

\section{ Diverse Redundant Safety Mechanisms}

\begin{table*}[!h]
\centering
\caption{Comparison of methods and measures for error detection of AI-based Software Elements}
\label{Tab:OOD}
\footnotesize
\fontsize{6pt}{5pt}\selectfont
\begin{tabularx}{\textwidth}{|X|X|X|X|X|X|X|}
\hline
\textbf{Method} & \textbf{Safety Measure/Mechanism Implementation Details} & \textbf{Pros} & \textbf{Cons} & \textbf{Is Run-time Measure?} & \textbf{Does the Run-time measure triggers the AI Element Architecture modification?} & \textbf{Can be a part of Majority Voter?} \\
\hline
Out of Distribution Detection & Reject Class & 1) No/less post-processing necessary (due to additional output class);2) Output conducive to safety-critical applications
3) Better performance of inputs that are near the boundary of training distribution; 4)Unlabeled data for OOD detection can be used if self-supervised learning is involved & 1)Prior knowledge of outliers or assumptions of outliers need to be made; 2)Training process needs to be altered; 3) Requires label OOD data (if not self-supervised); 4) Likely compute-time and memory overhead & Yes & AI Element Architecture & Yes \\
\hline
  & Isolation Forest & 1) Speed and efficiency;
2) Ability to handle high dimensional data;
3) Robustness towards outlier and noise;
4) Don’t require OOD data or class labels for ID data;
5) Can effectively handle global outliers & 1) Potential for False positive affecting the performance metric; 2) weak in dealing with local outliers; 3)Overhead of computation power and resources; 4) Outlier score doesn't improve interpretability (reasoning for outlier is not known) & Yes & AI Element Architecture & Yes \\
\hline
  & Local Outlier Factor &1)Can detect global and local outliers effectively;2)More effective detecting outliers that can't be detected by using global outlier detectors;3) No prior knowledge about the data distribution &1)Overhead of computation power and resources; 2)Outlier score doesn't improve interpretability (reasoning for outlier is not known) & Yes & AI Element Architecture & Yes \\
\hline
  & Softmax Function & 1)No compute time overhead;2)Inherent to large set of AI elements;3) Modification to AI architecture not necessary & Potential for false negatives/overconfidence & No & None & No \\
\hline
  & Temperature Scaling & 1) No compute time overhead; 2) Inherent to a large set of AI elements; 3) Modification to AI architecture and training process not necessary; 4) No prior knowledge of outliers or no assumptions of outliers need to be made  & 1) Tuning the temperature parameter in necessary due to sensitivity.; 2) Selective calibration for specific classes not feasible;3) Less effective in case of overlapping class distributions; 4) Potential for false negatives/overconfidence & No & None & No \\
\hline
  & Mahalanobis Distance & 1)Can detect outliers; 2) Used for multivariate data &1) Inverse of correlation matrix is needed; 2) Likely compute-time and memory overhead & No & None & No \\
\hline
Uncertainty Estimation & MC Dropout & 1) Uncertainty reflected in output; 2) Robustness to input variation & 1) Large computing power; 2) Large memory size & May be & AI Element Architecture & May be \\
\hline
  & Bayesian NN & 1) Uncertainty reflected in output; 2) Robustness to input variation/noise & 1) Large computing power; 2) Large memory size; 3) Increase in training time & May be & AI Element Architecture & May be \\
\hline
  & Deep Ensemble & 1) Uncertainty reflected in output; 2) Robustness to input variation/noise; 3) Diversity & 1) Large computing power; 2) Large memory size; & May be & AI Element Architecture & May be \\
\hline
\end{tabularx}
\end{table*}

The safety and reliability of AI in autonomous driving is a challenging topic, particularly when the system encounters scenarios outside its training distribution. To mitigate the risks associated with inputs out of training data sets, which may lead to overconfident and potentially hazardous decisions, a diverse redundancy approach can be employed. This involves integrating various error detection methods such as reject classes, monitoring based on IF and LOF, and possibly uncertainty estimation methods followed by specific post-processing to create a composite safety mechanism.

By adopting a majority voting system, these methods can be synthesized to enhance decision-making. For example, if two methods suggest an input is in-distribution but one strongly flags it as OOD, the system may default to treating the input as OOD, triggering cautious system behavior or requesting human intervention. This strategy is similar to sensor fusion in autonomous vehicles, where diverse sensors ensure robust perception against individual sensor errors.

The combination of different OOD detection techniques ensures that the AI system is not solely dependent on one method, thereby increasing its robustness. It also allows for the system to maintain operational safety by responding conservatively to uncertain data. As these methods detect and handle OOD events, they collectively inform system updates and training, enhancing the vehicle's capability to deal with a wider range of driving conditions. Ultimately, this multifaceted approach to OOD detection underscores a commitment to advancing autonomous vehicle safety, ensuring that AI systems remain reliable amidst the unpredictability of real-world driving.

\begin{figure}[!h]
\centering
\includegraphics[width=0.5\textwidth]{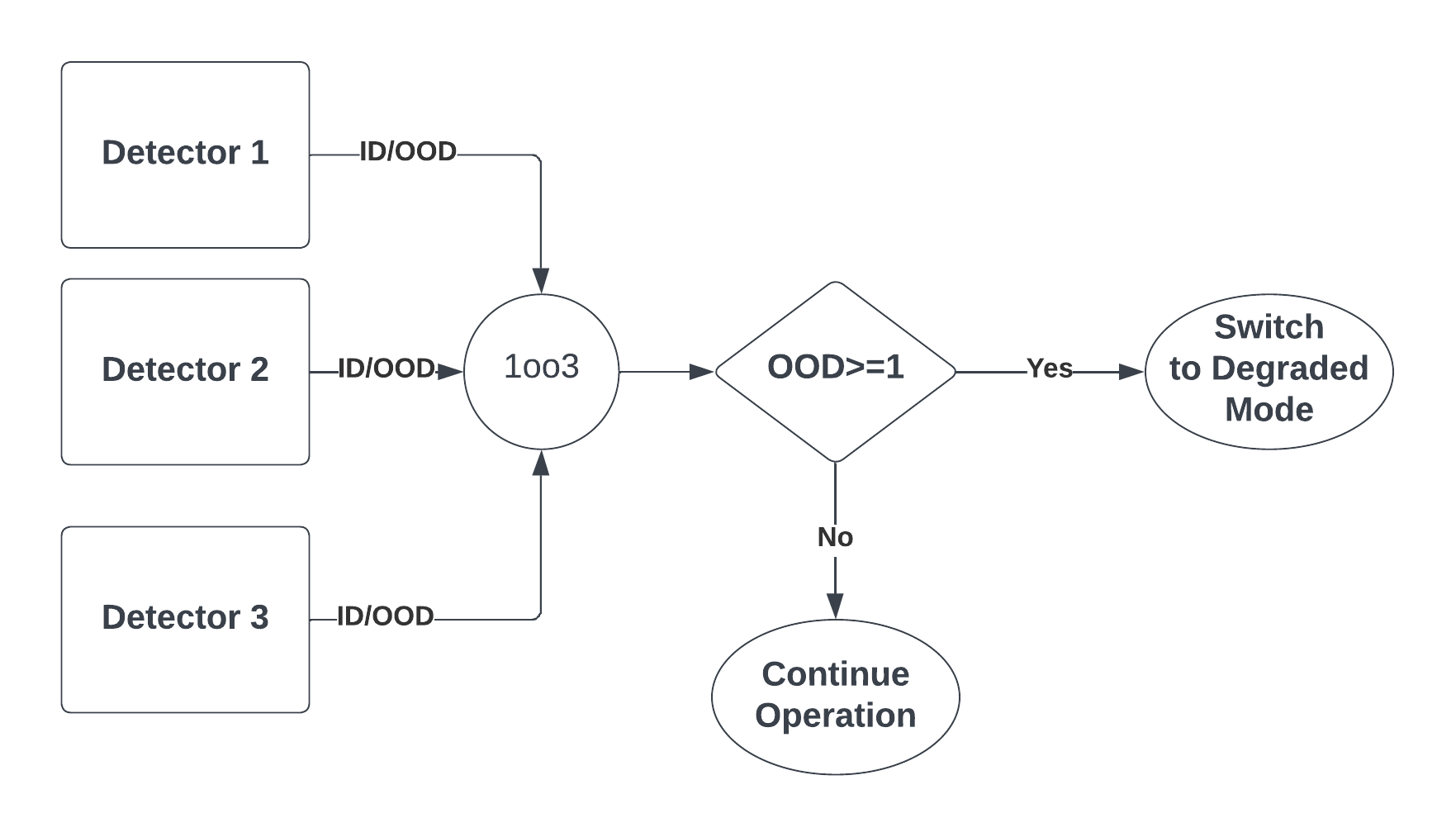}
\caption{ 1 out of 3 (1oo3) reliability checker block diagram}
\label{fig:reliable}
\end{figure}

Figure \ref{fig:reliable} shows the 1 out of 3 (1oo3) safety mechanism for detecting AI degradation. The design features a parallel three-channel configuration linked to a decision-making mechanism for the resultant signal. This mechanism adheres to a 1oo3 voting protocol, whereby the final state indicates the system's degradation when one or more of these three detectors report an error in the AI system.  \\

\begin{figure}[!h]
\centering
\includegraphics[width=0.5\textwidth]{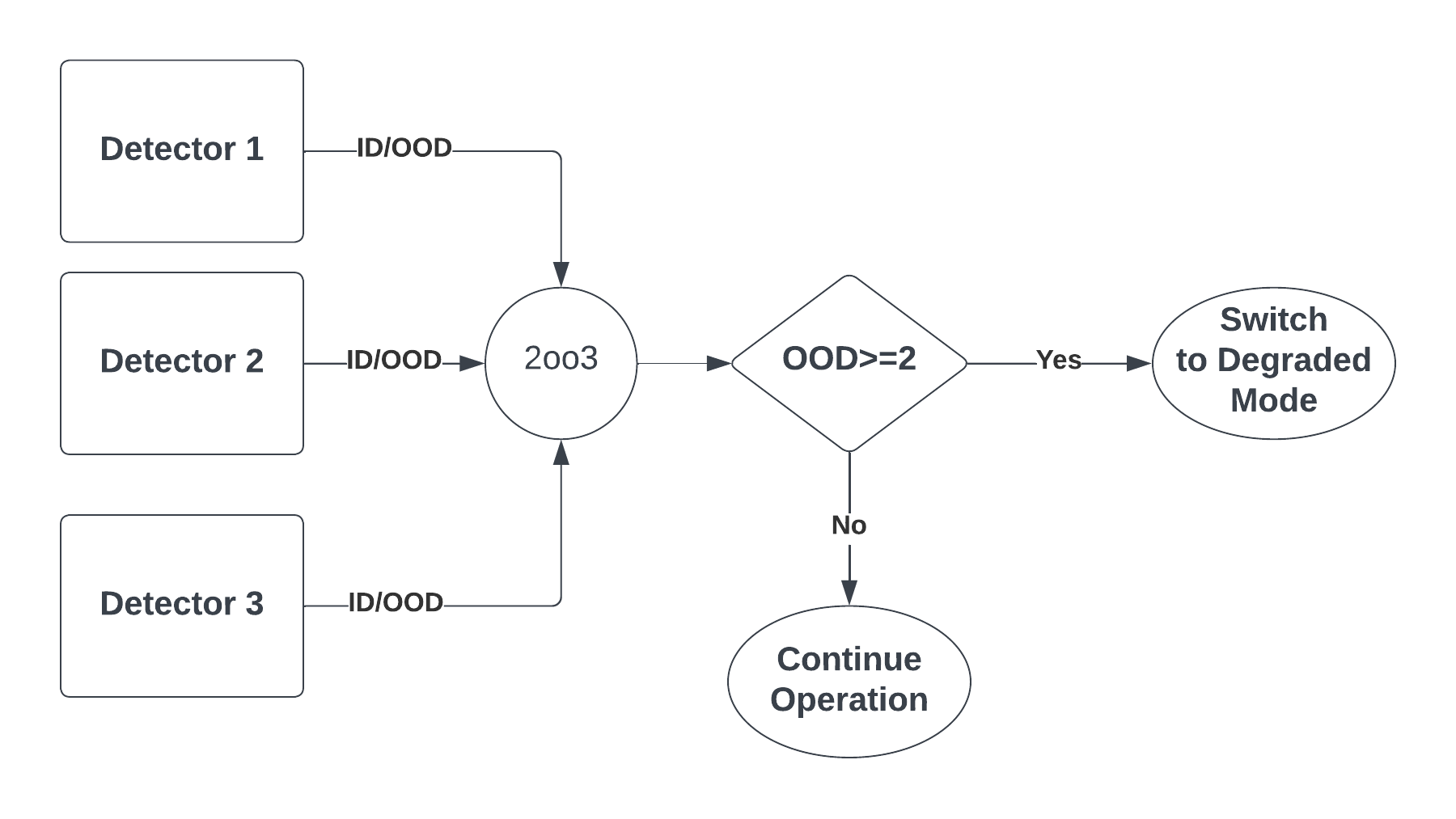}
\caption{ Majority voter of 2 out of 3 (2oo3) reliability checker block diagram}
\label{fig:Majority}
\end{figure}

Figure \ref{fig:Majority} illustrates the 2 out of 3 (2oo3) safety mechanisms employed to identify AI degradation. The architecture is designed with a tri-channel arrangement operating in parallel, converging on a decision logic module that adjudicates the AI system's integrity. The mechanism operates on a 2oo3 voting protocol, requiring at least two of the three detectors to concurrently report an anomaly before confirming the system's degradation. 

The 1oo3 reliability checker and the 2oo3 majority voter systems embody a trade-off strategy between the risks of false positive (FP) and false negative (FN) detection errors, specifically addressing OOD detection in autonomous driving safety systems. A 1oo3 system indicates an OOD occurrence if any of the detectors report it, which minimizes the chance of missing an actual OOD event (lower FN) but raises the probability of indicating an OOD event when there is none (higher FP). This system adheres to a 'fail-safe' philosophy, prioritizing conservative safety actions over operational continuity. On the contrary, a 2oo3 majority voter system demands a majority agreement to confirm an OOD event, effectively reducing FP incidents but at the risk of a higher FN rate due to possibly ignoring genuine OOD events not concurrently acknowledged by two detectors. A lower FP rate is beneficial for avoiding unnecessary disruptions and maintaining system efficiency, while a higher FN rate may be permissible, provided the undetected OOD events do not critically undermine system safety. Choosing between these systems requires careful consideration of the permissible risk profiles for FP and FN within the safety management strategy of autonomous driving systems. The 1oo3 scheme is suited to contexts where safety takes precedence, and the impact of FN on system performance is unacceptable, whereas the 2oo3 configuration may be optimal where the implications of FP are substantial, and the FN risk is considered to be within acceptable limits. Table \ref{table:methods} summarizes the pros and cons of the 1oo2 and 2oo3 reliability checkers. Based on the criticality and priority of the situation between FP and FN, the system designer can choose between these methods. Also, an AI algorithm could be trained to decide when to use 1oo3 or 2oo3 in different situations.

\begin{table}[h]
\centering
\caption{Comparison of voting mechanism methods based on false negatives (FN) and false positives (FP).}
\begin{tabular}{|c|c|c|}
\hline
\textbf{Method} & \textbf{Advantage} & \textbf{Limitation} \\ \hline
1 out of 3      & Low FN             & High FP             \\ \hline
2 out of 3      & Low FP             & High FN             \\ \hline
\end{tabular}
\label{table:methods}
\end{table}

\subsection{Example design of the Voter}

\begin{figure}[!h]
\centering
\includegraphics[width=0.5\textwidth]{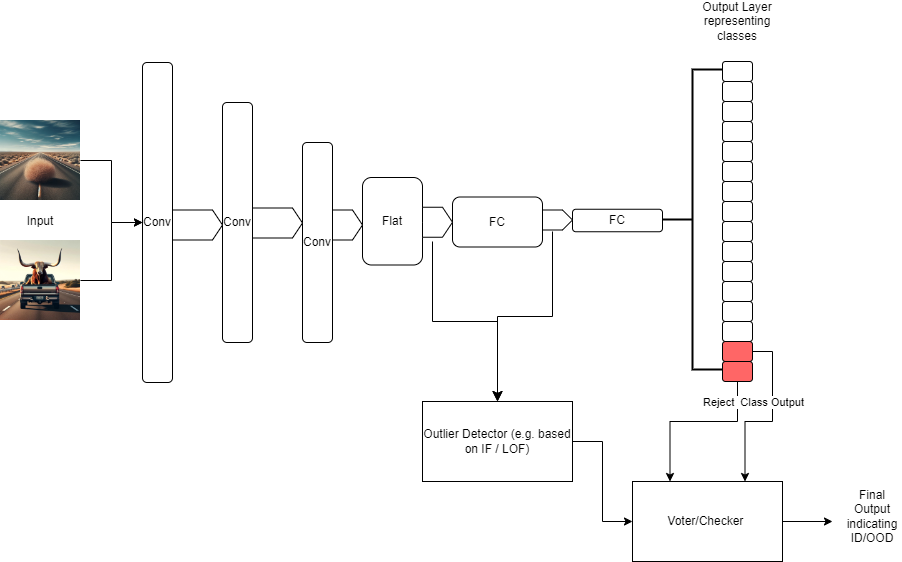}
\caption{Inherent Diverse Safety Mechanism.}
\label{fig:diverse}
\end{figure}

Fig. \ref{fig:diverse} shows the overall block diagram of the proposed inherent diverse redundant safety mechanism for AI models with the application of a voter. Typically for a simple voter, the redundant safety mechanisms need to provide output that can be easily compared without adding significant post-processing or conversion overhead that impacts the compute time and resources. As can be seen in the figure the intermediate layers of the AI element are used as an input of the monitor based on IF or LOF \cite{luan2021out} (or based on any other comparable measure). The other inputs are from the reject classes \cite{mohseni2020self}. In such a setting, as the outputs are typically binary it is relatively simple to design a voter using 1oo3 or 2oo3 reliability checker pattern as described in the previous section.   

\section{ Conclusion}
In this paper, we highlighted the problem of the overconfident AI model and then reviewed the state-of-the-art inherent safety mechanisms for AI-based software elements. We also identified if these safety mechanisms can be essentially used as a run-time measure. We also presented a new method to use multiple redundant safety mechanisms targeting their diversity in principle and implementation and to be used as an input to the voter. In the end, the introduction of a voter was to improve the safety and reliability of the overall error detection method and also explain how a specific configuration of the voter can affect the performance metrics (e.g., FP, FN) of AI elements.   

\subsection{Future Work}
The majority of the discussed and leveraged safety mechanisms are focused on classifiers; however, in future work, we will review how the presented solution can be applied to a regression-based AI model. For In-Distribution Detection, Distribution shift Detection, and Adversarial perturbation detection, we plan to investigate what safety mechanisms can be employed as run-time measures. Furthermore, a detailed study of the diversity claims about these redundant safety mechanisms is also planned to be done in the next step. Importantly, we plan to perform experiments on these diverse redundant safety mechanisms to identify how their impact on the performance influences the performance metric.


\bibliographystyle{ieeetr} 
\bibliography{AI_safety}
\end{document}